\documentclass[10pt,twocolumn,letterpaper]{article}

\usepackage{cvpr}
\usepackage{times}
\usepackage{epsfig}
\usepackage{graphicx}
\usepackage{amsmath}
\usepackage{amssymb}
\usepackage[dvipsnames]{xcolor}

\newcommand{\red}{\textcolor{black}}

\usepackage[pagebackref=true,breaklinks=true,letterpaper=true,colorlinks,bookmarks=false]{hyperref}

\cvprfinalcopy 

\def\cvprPaperID{03} 

\begin{document}

\title{\vspace{-7mm}Cleaning Label Noise with Clusters for Minimally Supervised Anomaly Detection}

\author{Muhammad Zaigham Zaheer$^{1,2}$, Jin-ha Lee$^{1,2}$, Marcella Astrid$^{1,2}$, Arif Mahmood$^{3}$, Seung-Ik Lee$^{1,2}$\\
$^{1}$University of Science and Technology, $^{2}$Electronics and Telecommunications Research Institute,\\
Daejeon, South Korea.\\
$^{3}$Information Technology University, Lahore, Pakistan\\
{\tt\small \{mzz, jhlee, marcella.astrid\}@ust.ac.kr,}
{\tt\small  arif.mahmood@itu.edu.pk,}
{\tt\small  the\_silee@etri.re.kr}
}
\maketitle
\thispagestyle{empty}

\begin{abstract}

Learning to detect real-world anomalous events using video-level annotations is a difficult task mainly because of the noise present in labels. An anomalous labelled video may actually contain anomaly only in a short duration while the rest of the video can be normal.
In the current work, we formulate a weakly supervised anomaly detection method that is trained using only video-level labels. To this end, we propose to utilize binary clustering which helps in mitigating the noise present in the labels of anomalous videos. Our formulation encourages both the main network and the clustering to complement each other in achieving the goal of weakly supervised training.
The proposed method yields 78.27\% and 84.16\% frame-level AUC on UCF-crime and ShanghaiTech datasets respectively, demonstrating its superiority over existing state-of-the-art algorithms.
\end{abstract}

\section{Introduction}
Anomalous event detection is a challenging problem in computer vision because of its applications in real-world surveillance systems \cite{sultani2018real}. 
Due to infrequent occurrences, anomalous events are usually seen as outliers from the normal behavior \cite{zaheer2020old}. Hence, anomaly detection is often carried out using one-class classifiers which learn the commonly occurring events as normal. Anomalies are then detected based on their deviations from the learned representations. However, it is not always feasible to collect every possible normal scenario for training, therefore new occurrences of normal events may also substantially differ from the learned representations and may be detected as anomalous.

Another anomaly detection approach is to utilize the weakly supervised learning paradigm to train a binary classifier using both normal and anomalous data instances \cite{sultani2018real,zhong2019graph}. In such setting, presence of all normal events in a video is marked as normal whereas presence of some anomalous scenes in an otherwise normal video is marked as anomalous. Though, it reduces the efforts required in obtaining detailed manual annotations of the dataset, the training using this type of labels is quite challenging.
In the current paper, we propose an approach for anomalous event detection using such video level labels.

\begin{figure*}[t]
\begin{center}
\includegraphics[width=0.85\linewidth]{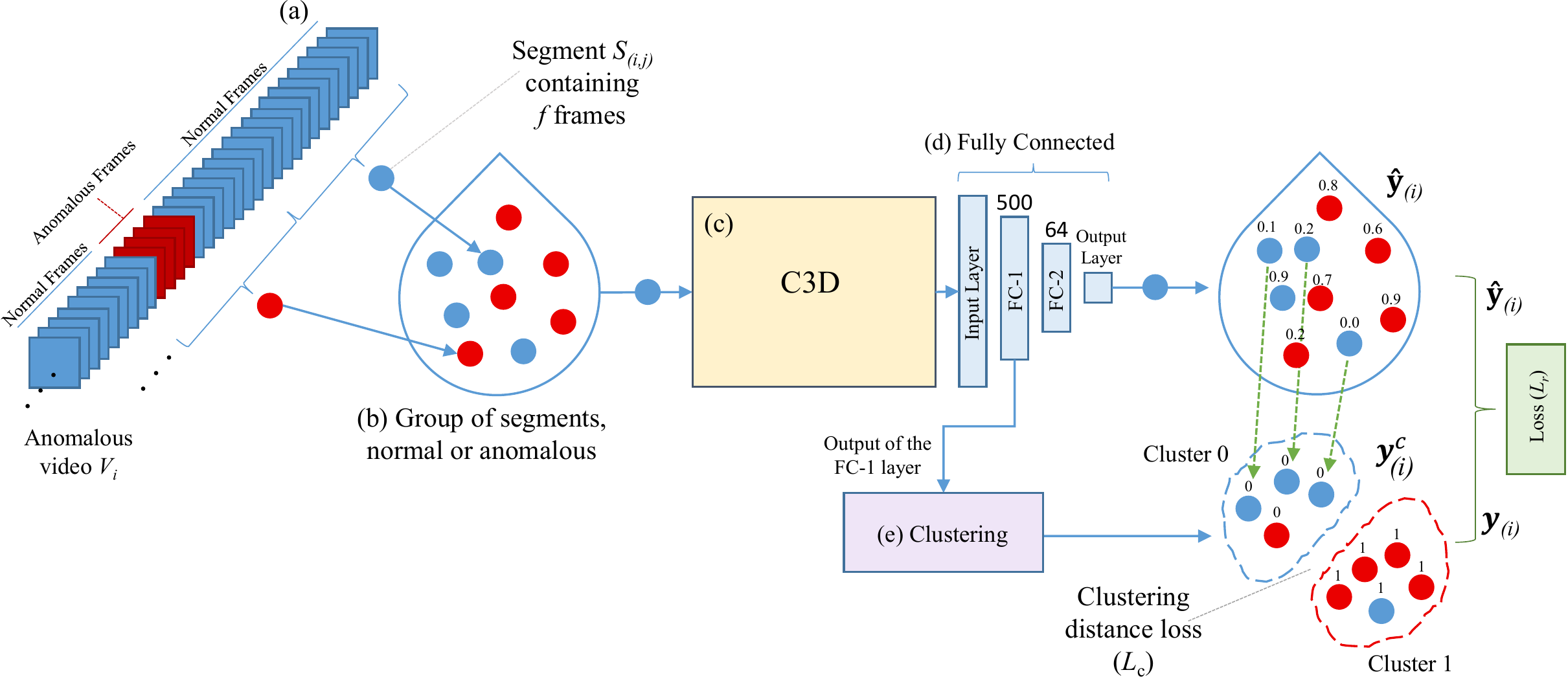}
\end{center}
  \caption{Our proposed architecture for anomaly detection in weakly supervised setting. The labels are provided only at video-level. The video frames (a) are converted into a group of segments (b). Feature extraction is performed on each segment (c) and the features are input to the fully connected network (d). Intermediate representations of a whole video inferred from the FC-1 layer are used to create clusters (e). For an anomalous labelled video, the pseudo labels $y^p$ are generated with the help of clusters.}
\label{fig:architecture}
\end{figure*}

Recently, anomaly detection problem in weakly labelled videos has been formulated as Multiple Instance Learning (MIL) task \cite{sultani2018real}. A bag of segments is created using one complete video in such a way that each segment consists of several consecutive frames of the video. Training of the network is then carried out by defining a ranking loss between two top-scoring segments, each from an anomalous and a normal bag.
However, this approach necessitates to compress each video of the dataset into the same number of segments which is not always appropriate. 
Since the real-world datasets contain significantly varying length of videos, a rigid formulation may not be able to represent events happening over a short span of time.
Zhong \etal \cite{zhong2019graph} has also proposed an anomaly detection approach using weakly labelled videos. In their approach, training is performed using noisy labels, where the noise refers to normal segments within anomalous videos. 
They take advantage of an action recognition model to train a graph convolution network, which then helps in cleaning noisy labels from the anomalous videos. In essence, our approach is similar to theirs because we also attempt to eliminate noisy labels. However, the way we formulate the problem is completely different. In our architecture, instead of graph convolution network, we propose to employ a binary clustering based approach which not only attempts to remove noisy labels but also contributes in enhancing the performance of our model through a clusters distance loss. 
The main contributions of our work are summarized below:

\begin{itemize}
  \item Our proposed framework trains in a weakly supervised manner using only video-level annotations to detect anomalous events.
 \item We propose to employ clustering to clean noise from the labels of anomalous videos. Our framework allows the learning network to enhance clusters over time, hence enabling both the network and the clustering algorithm to complement each other during training.
  \item The framework demonstrates state-of-the-art results by yielding frame level AUC performance of \red{78.27\%} on UCF-crime \red{\cite{sultani2018real}} dataset and \red{84.16\%} on ShanghaiTech \cite{luo2017shanghaitech} dataset.
\end{itemize}

\section{Proposed Architecture}
The overall framework is visualized in Figure \ref{fig:architecture}, whereas each of its components are discussed below:


\noindent\textbf{Group of Segments:}
All frames from a complete video $V_i$ are divided into a group of segments in such a way that each segment $S_{(i,j)}$ contains $f$ non-overlapping frames, where $i\in[1,n]$ is the video index in the dataset of $n$ videos and $j\in[1,m_i]$ is the index of $m_i$ segments in $V_i$. For each video, only binary labels \{normal = 0, anomalous = 1\} are provided.

\noindent\textbf{Feature Extractor:}
Features of each $S_{(i,j)}$ are computed by employing a pre-trained feature extractor model such as Convolution 3D (C3D) \cite{tran2015c3d}.

\noindent\textbf{Fully Connected Network:}
To learn the feature representations, two fully connected layers are employed, each followed by a ReLU activation function and a dropout layer. The input layer receives a feature vector and the output layer produces an anomaly regression score in the range of $[0, 1]$ through a sigmoid activation function.

\noindent\textbf{Clustering:}
Given that anomaly detection is a binary problem, clustering algorithms such as k-means \cite{kanungo2002kmeans} can be employed to distribute all segments into two clusters. These clusters are created using the feature representations of each segment taken from the output of FC-1 layer.
Clustering here serves two purposes: 1) it helps in generating segment-level pseudo annotations from video-level labels. 2) It encourages the network to push both clusters away in the case of an anomalous video, and brings both clusters closer in the case of a normal video.

\subsection{Training}
As explained previously, our architecture is trained using only video-level labels, therefore we utilize clustering to create pseudo annotations. Moreover, the configuration also encourages fully connected network and clustering to complement each other towards improving the results over training iterations.

\noindent\textbf{Creating Pseudo Annotations:}
For the normal labelled videos, as no anomaly is present, each segment of these videos can be annotated as normal. However, in the case of anomalous videos, several normal segments may also be present. Therefore, pseudo annotations are only generated for anomalous videos. To this end, all segments from an anomalous video are divided into two clusters assuming one cluster should contain normal while the other should contain anomalous segments. To determine pseudo annotation $y^p_{(i,j)}$, the similarity between predicted scores and cluster labels is computed. This way, we rely on the experience of FC network to identify normal segments which it obtains through noise-free labels and find out the cluster containing most of the normal segments.
Given $j^{th}$ segment in $V_i$, $y^p_{(i,j)}$ is defined as:
\begin{equation}
    y^p_{(i,j)}=
    \begin{cases}
    y^c_{(i,j)}, & \text{if } s_1 \geq s_2\\
    1 - y^c_{(i,j)}, & otherwise, 
    \end{cases}
\end{equation}
where clustering labels $y^c_{(i,j)} \in \{0,1\}$ depending on the placement of $S_{(i,j)}$ in either of the clusters. Furthermore, $s_1$ and $s_2$ are given as:
\begin{equation}
s_1 =  \frac{\hat{\mathbf{y}}_{(i)}.\mathbf{y}^c_{(i)}}{||\hat{\mathbf{y}}_{(i)}||_2\times||\mathbf{y}^c_{(i)}||_2} ,
\end{equation}
\begin{equation}
s_2 = \frac{\hat{\mathbf{y}}_{(i)} . (\mathbf{1} - \mathbf{y}^c_{(i)})}{||\hat{\mathbf{y}}_{(i)}||_2\times||\mathbf{1}-\mathbf{y}^c_{(i)}||_2} ,
\end{equation}
where $\hat{\mathbf{y}}_{(i)} \in [0,1]^{m_i}$ is a vector containing prediction scores for all segments of $V_i$ and $\mathbf{y}^c_{(i)} \in \{0,1\}^{m_i}$ is a vector containing clustering labels for all segments of $V_i$.
Finally, the segment-level annotations for training are given as:
\begin{equation}
    y_{(i,j)}=
    \begin{cases}
    0, & \text{if $V_i$ is normal}\\
    y^p_{(i,j)}, & \text{if $V_i$ is anomalous.} 
    \end{cases}
    \label{eq:labels}
\end{equation}
\subsection{Training Losses}
Overall, our network is trained to minimize the loss: 
\begin{equation}
    L= L_r + \lambda L_c,
\end{equation}
where $L_r = MSE (y_{(i,j)}, \hat{y}_{(i,j)})$. Moreover, clustering distance loss, $L_c$, is defined as:
\begin{equation}
    L_c=
    \begin{cases}
    min(\alpha, d_i), & \text{if $V_i$ is normal}\\
    \frac{1}{d_i}, & \text{if $V_i$ is anomalous,} 
    \end{cases}
\end{equation}
where $d_i$ is the distance between the centers of the two clusters formed using the segments of $V_i$, $\alpha$ is an upper bound on the distance loss and $\lambda$ is a trade-off hyperparameter.

\section{Experiments}
\subsection{Datasets}
\noindent\textbf{UCF-crime} \cite{sultani2018real}: It is a weakly labelled abnormal event dataset obtained from real-world surveillance videos. For training, it contains 810 videos of anomalous and 800 of normal classes. For testing, it contains 140 anomalous and 150 normal videos.

\noindent\textbf{ShanghaiTech} \cite{luo2017shanghaitech}: It is also an anomalous event dataset recorded in a university campus. The original training split of this dataset does not contain any anomalous videos because it follows one class classification protocol. However Zhong \etal\cite{zhong2019graph} proposed a split which contains 63 anomalous and 175 normal videos for training and 44 anomalous and 155 normal videos for testing which can be used for binary learning algorithms such as ours.

For both datasets, Area Under the Curve (AUC) of the Receiver Operating Characteristic (ROC) curve for frame-level performance is computed as the evaluation metric. Adam optimizer is used with a learning rate of 5 $\times$ $10^{-5}$ whereas, $\alpha$ and $\lambda$ are set to 1 and 0.05 respectively. Unlike some of the compared models that use several types of feature extractors, we employ only C3D architecture \cite{tran2015c3d}. Therefore, for fair comparison, we report results only with the C3D features. Moreover we use default settings of C3D in which $f$, the number of frames per segment, is set to 16.
\subsection{Results}
\noindent\textbf{UCF-crime:}
Table \ref{tab:UCF_crime_AUC} summarizes a comparison of our approach with the existing state-of-the-art methods. Our approach outperforms most of the compared algorithms with a significant margin. The method proposed in Zhong \etal \cite{zhong2019graph} performs better, however the results are comparable.

\noindent\textbf{ShanghaiTech:}
As mentioned previously, the split of this dataset for weakly supervised learning was recently introduced by Zhong \etal \cite{zhong2019graph}. Results presented in Table \ref{tab:shanghaitechAUC} show that using the same dataset protocol and C3D features our algorithm outperforms Zhong \etal \cite{zhong2019graph} by a significant margin of \red{7.72\%} in AUC. 

\noindent\textbf{Ablation Study:}
A detailed ablation study on both UCF Crime and ShanghaiTech Datasets is provided in Table \ref{tab:ablation}. We follow a top-down approach in which several components of the network are individually removed to observe their significance. In the case of ShanghaiTech dataset, removal of clustering distance loss ($L_c$) resulted in a drop of \red{0.79\%} whereas removal of clustering based pseudo annotations $\mathbf{y}^p$ for anomalous videos resulted in a drop of \red{2.51\%}. Experiments on UCF-crime dataset also demonstrated similar trends in which removal of $L_c$ resulted in a drop of \red{1.09\%} whereas removal of $\mathbf{y}^p$ resulted in a drop of \red{1.63\%}. Note that the labels of all segments of an anomalous videos are set to 1 in Equation \ref{eq:labels} when we remove $\mathbf{y}^p$.

\begin{table}[t]
\begin{center}
\begin{tabular}{c|c} 
\text{\textbf{Method}} & \text{\textbf{AUC(\%)}} \\ \hline
Binary SVM \cite{sultani2018real}   & 50.00\\ \hline
Hasan \etal \cite{hasan2016anomaly}   & 50.60\\ \hline
Lu  \etal \cite{lu2013abnormal}     & 65.51\\ \hline
Sultani  \etal \cite{sultani2018real}   & 75.41\\ \hline
Zhong  \etal \cite{zhong2019graph} &  \textbf{81.08}\\ \hline\hline
\textbf{Ours} & \red{\underline{78.27}}
\end{tabular}
\end{center}
\caption{UCF-crime Dataset: Frame-level AUC performance comparison of our approach with state-of-the-art methods. Best performance marked as bold, second best marked as underlined.}
\label{tab:UCF_crime_AUC}
\end{table}

\begin{table}[t]
\begin{center}
\begin{tabular}{c|c}
\textbf{Method} & \textbf{AUC \%} \\ \hline
Zhong et al. \cite{zhong2019graph} & \underline{76.44}  \\ \hline\hline
\textbf{Ours} & \textbf{\red{84.16}}
\end{tabular}
\end{center}
\caption{ShanghaiTech Dataset: Frame-level AUC performance comparison of our approach with state-of-the-art methods using C3D features. Best performance marked as bold, second best marked as underlined.}
\label{tab:shanghaitechAUC}
\end{table}

\begin{table}[t]
\begin{center}
\begin{tabular}{c|c|c}
                           & \multicolumn{2}{c}{\textbf{Datasets}} \\ \hline
\textbf{Method}                  & \textbf{Shanghai Tech}   & \textbf{UCF-crime}   \\ \hline
FC + $L_c$ + $y^p$              & \red{84.16}           & \red{78.27}  \\ \hline
FC + $y^p$  & \red{83.37}            & \red{77.13}        \\ \hline
FC + $L_c$         & \red{81.65}            & \red{76.59}        \\ 

\end{tabular}
\end{center}
\caption{Top-down ablation of our approach provided on ShanghaiTech and UCF-crime datasets.}
\label{tab:ablation}
\end{table}

\begin{figure*}[t]
\begin{center}
\includegraphics[width=.84\linewidth]{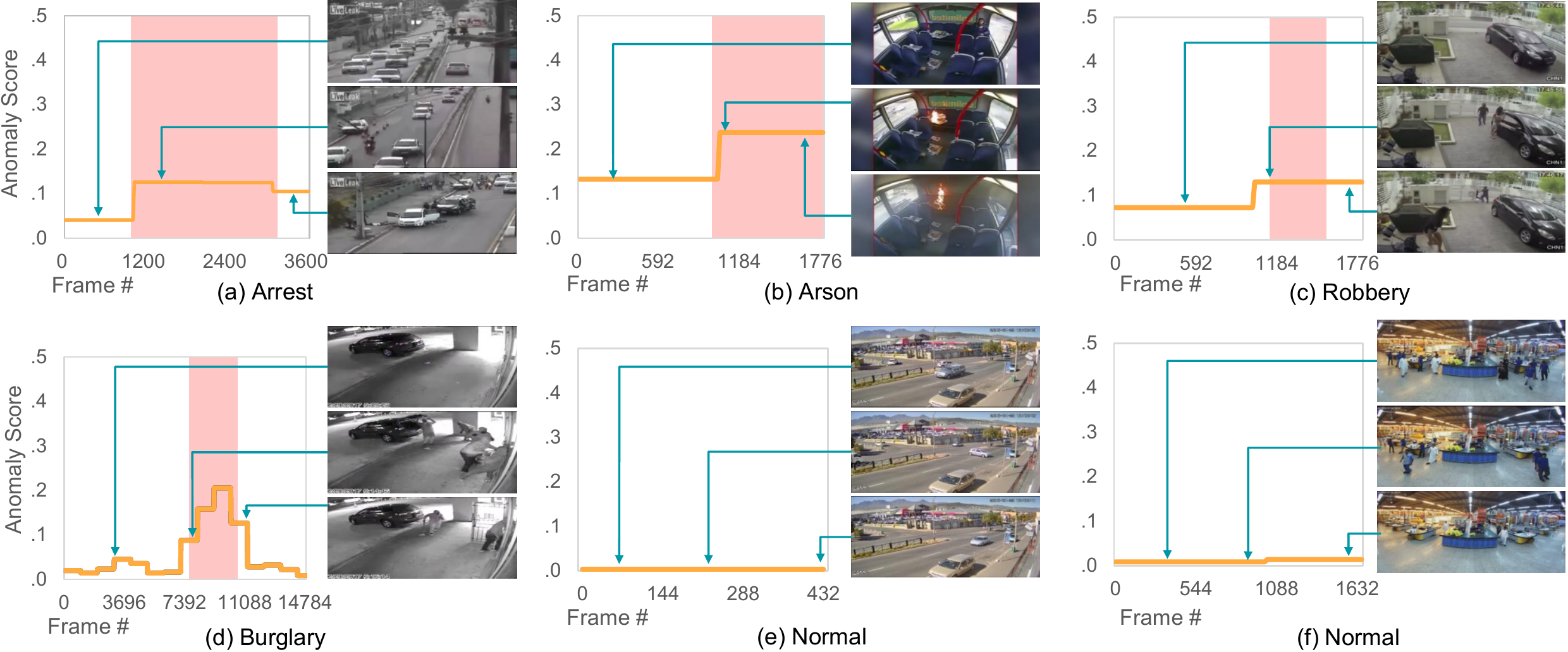}
\end{center}
  \caption{Qualitative results of our approach on test videos of UCF-crime dataset. Colored rectangular window represents anomaly ground truth. (a),(b),(c), and (d) show the scores predicted by our model on anomalous whereas (e) and (f) show the scores on normal videos.}
\label{fig:qualitative}
\end{figure*}

\noindent\textbf{Qualitative Results:}
Anomaly score plots of several normal and anomalous test videos from the UCF-crime dataset are visualized in Figure \ref{fig:qualitative}. Overall, our network produces distinctive scores for anomalous portions of the videos.
\vspace{-2mm}
\section{Conclusion}
\vspace{-1mm}
In this paper, a weakly supervised learning approach using video-level labels to detect anomalous events is proposed. 
Compared to using frame level annotations, the video level annotations contain significant noise.
It is because an anomalous labelled video may contain anomaly only in a short duration while the rest of the video may be normal.
Therefore training using such noisy labels is a very challenging task. To this end, binary clustering is employed to mitigate the noise present in the labels of anomalous videos. The proposed framework enables both the fully connected network and the clustering algorithm to complement each other in improving the quality of results. Our method demonstrates state-of-the-art results by yielding \red{78.27\%}  and  \red{84.16\%} frame-level AUC performances on the UCF-crime and ShanghaiTech datasets respectively.

\vspace{-2mm}
\section{Acknowledgment}
\vspace{-2mm}
This work was supported by the ICT R\&D program of MSIP/IITP. [2017-0-00306, Development of Multimodal Sensor-based Intelligent Systems for Outdoor Surveillance Robots].

\vspace{-3mm}
{\small
\bibliographystyle{ieee_fullname}
\bibliography{egbib}
}

\end{document}